%% file: main.tex
\documentclass[10pt,twocolumn,letterpaper]{article}

\usepackage[pagenumbers]{cvpr} 

\input{preamble}

\definecolor{cvprblue}{rgb}{0.21,0.49,0.74}
\usepackage[pagebackref,breaklinks,colorlinks,allcolors=cvprblue]{hyperref}
\usepackage{fontawesome5}
\def\confName{CVPR}
\def\confYear{2027}
\usepackage{etoolbox}
\makeatletter
\patchcmd{\@maketitle}
  {\vspace*{24pt}}
  {\vspace*{10pt}}
  {}{}

\patchcmd{\@maketitle}
  {\vskip .375in}
  {\vskip .15in}
  {}{}
\makeatother
\title{ReAccess: Look Once, Stay Focused with Efficient Test Time Memory}

\author{
Tianhao Chen$^{1}$ \quad
Yilong Fan$^{2}$ \quad
Yuheng Wu$^{1}$ \quad
Xiaobin Hu$^{3,*}$ \quad
Dongman Lee$^{1,*}$ \\[0.3em]
{
$^{1}$KAIST \qquad
$^{2}$Tianjin University \qquad
$^{3}$National University of Singapore
}\\[0.5em]
{
\href{https://ryu1ion.github.io/reaccess/}{\faGlobe\ \textbf{Project Page}}
\qquad  \qquad
\href{https://github.com/ryu1ion/reaccess}{\faGithub\ \textbf{Code}}
}
}

\begin{document}
\maketitle
\begingroup
\renewcommand{\thefootnote}{\fnsymbol{footnote}}
\footnotetext[1]{Co-corresponding authors.}
\endgroup
\input{sec/0_abstract}

\input{sec/1_intro}

\input{sec/2_related}

\input{sec/3_motivation}

\input{sec/4_method}

\input{sec/5_exp}

\section*{LLM Usage Statement}
We used ChatGPT (OpenAI) to assist with code implementation and language polishing of the manuscript, including improvements to grammar, clarity, and readability. All AI-assisted content was reviewed and refined by the authors, who take full responsibility for the final implementation and scientific content.

{
    \small
    \bibliographystyle{ieeenat_fullname}
    \bibliography{main}
}

\input{sec/X_suppl}
\end{document}

%% file: preamble.tex
\usepackage{microtype}

\usepackage{booktabs,tabularx,array,graphicx,xcolor,colortbl,multirow,pifont}
\usepackage[most]{tcolorbox}

%% file: sec/0_abstract.tex
\begin{abstract}
Despite recent advances, vision-language models (VLMs) still struggle with fine-grained visual understanding, often failing to effectively utilize critical evidence already encoded in their visual representations. Existing approaches address this limitation through repeated visual processing or explicit reasoning, introducing substantial computational overhead. In this paper, we propose ReAccess, a lightweight framework that preserves and reuses question-relevant visual evidence through compact associative memory. Constructed during a single visual prefill, the memory supports progressive evidence retrieval that steers both attention queries and hidden representations, enabling effective visual grounding without repeated image encoding or explicit reasoning traces. Extensive experiments demonstrate consistent improvements in both general visual understanding and fine-grained perception, while achieving substantially faster inference than existing reasoning-intensive approaches, demonstrating a promising path to efficient multimodal reasoning. 
\end{abstract}

%% file: sec/1_intro.tex
\section{Introduction}
\label{sec:introduction}

Vision-Language Models (VLMs) \cite{Qwen2.5-VL,liu2023llava} have made remarkable progress in visual understanding and multimodal reasoning, enabling increasingly capable systems to interpret complex scenes. Yet richer visual perception does not always translate into more reliable answers \cite{tong2024eyes, liu2026seeing}. While images contain numerous objects, attributes, and spatial relationships, answering a question often hinges on a few fine-grained clues interpreted in the context of both the question and the broader scene. Reliable visual understanding therefore requires more than encoding rich visual information: a model must identify task-relevant evidence inside complex visual scenes and effectively leverage it during reasoning. This highlights a fundamental challenge: \emph{bridging the gap between visual information encoded by the model and its effective utilization in evidence-grounded reasoning}.

\begin{figure*}[t]
    \centering
    \includegraphics[width=\textwidth]{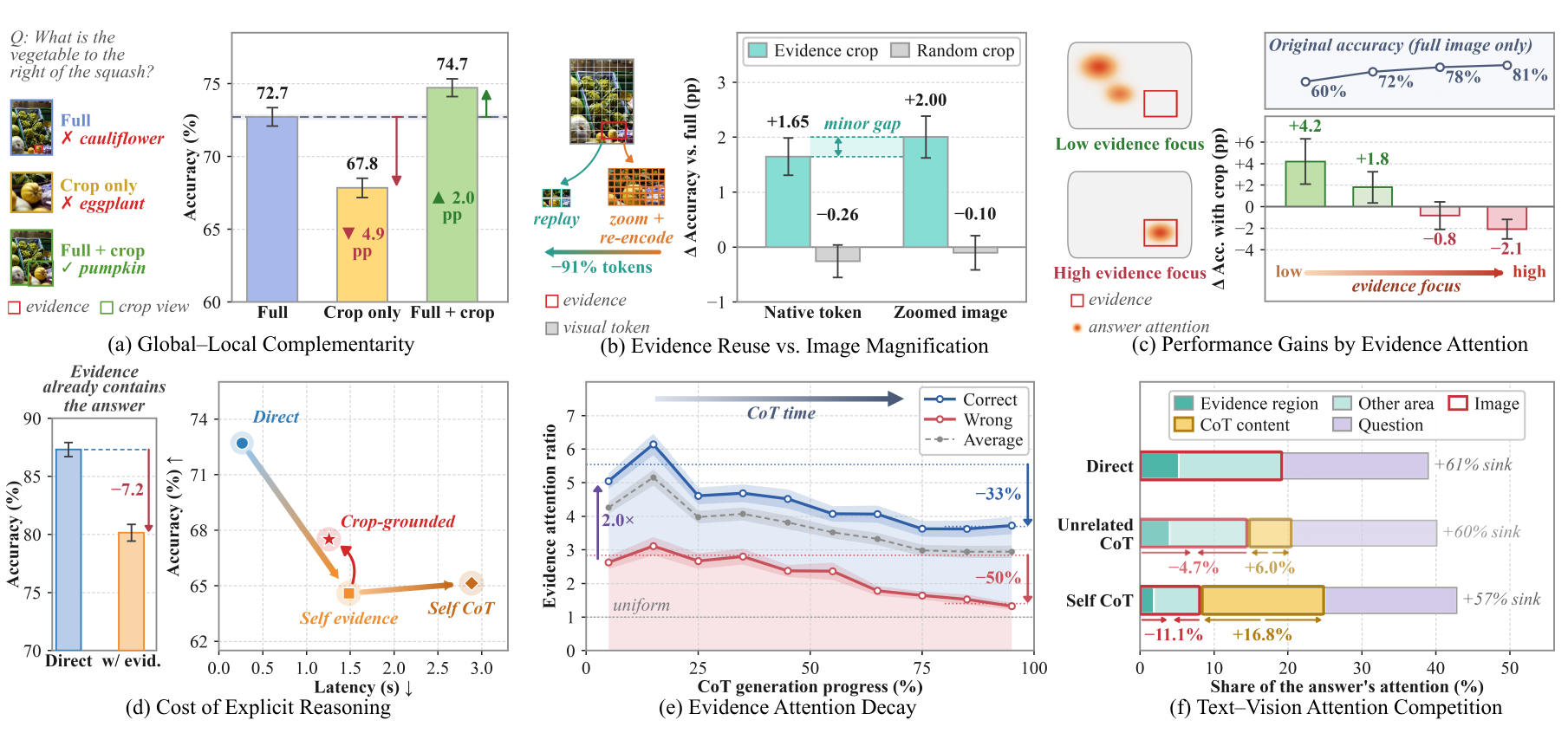}\par
    \vspace{-4mm}
    \caption{\textbf{Visual evidence accessibility in VLMs.}
    Analysis of focused visual evidence and its utilization during language reasoning.}
    \label{fig:motivation}
    \vspace{-6.5mm}
\end{figure*}

Prior work has explored two complementary directions to address this challenge. Visual methods \cite{wu2024v*, zheng2026deepeyes, zhang2025thyme} focus perception on informative regions, while language methods \cite{shao2024visual, wei2022chain} textualize observations through intermediate descriptions and reasoning traces. While both directions aim to improve evidence utilization, they introduce distinct trade-offs: additional visual observations increase computational overhead and may disrupt global context, whereas textualization can lose fine-grained details and spatial relations. These trade-offs motivate latent reasoning methods \cite{li2026latent,zou2024look}, which aim to preserve visual information without repeated image processing or textualization. Beyond these trade-offs, many methods require costly fine-tuning, potentially risking catastrophic forgetting \cite{forget}. More fundamentally, making richer visual information available does not necessarily translate into its effective utilization during reasoning. This reveals a more critical challenge: improving visual reasoning requires not only preserving relevant evidence, but selectively accessing and integrating it into reasoning.

Our analysis in \cref{fig:motivation,tab:motivation} provides further insight into this challenge. Focusing on informative regions can recover overlooked visual details, yet their interpretation still benefits from surrounding context. More importantly, as intermediate reasoning text accumulates, attention shifts away from the image, weakening the influence of visual evidence on predictions. Meanwhile, substantial visual information remains recoverable from highly compressed internal representations, suggesting the potential for compact evidence preservation. Together, these observations reveal that the central difficulty, further explored in \cref{sec:motivation}, lies not merely in preserving visual information, but in keeping relevant evidence accessible and influential throughout reasoning.

These findings motivate a more active role for visual memory. Rather than simply introducing additional representations, memory should selectively preserve question-relevant evidence in a compact form while maintaining access to the original visual context. More importantly, the stored evidence should be retrieved as the model's computation evolves, directly influencing how it attends to visual information and refines its predictions. The key is therefore to connect \emph{what evidence is preserved} with \emph{how that evidence is utilized}, transforming memory from passive storage into an active mechanism for visual reasoning. As evidence relevance depends on the current input and evolves throughout reasoning, such coordination calls for a mechanism that dynamically manages visual evidence during inference.

We introduce \textbf{ReAccess}, a lightweight \textbf{test-time evidence memory} framework that enables direct evidence utilization in frozen VLMs. During a single visual prefill, ReAccess constructs a compact, question-conditioned associative memory while preserving the original visual context. During reasoning, successive memory reads retrieve complementary visual evidence and integrate it into the model's native computation to guide predictions. By unifying selective evidence preservation with progressive utilization, ReAccess strengthens visual grounding without repeated image encoding or explicit intermediate reasoning.

Across diverse VLMs and benchmarks, ReAccess consistently improves visual understanding while achieving a better performance--efficiency trade-off than competing approaches. Our contributions are as follows:
\begin{itemize}
    \item We identify a gap between visual evidence representation and utilization, motivating sustained access to relevant evidence throughout reasoning.

    \item We propose \textbf{ReAccess}, a lightweight test-time memory unifying evidence preservation and adaptive utilization to enhance visual reasoning in frozen VLMs.

    \item We evaluate ReAccess across models and benchmarks against various baselines, showing consistent gains and a better performance--efficiency trade-off.
\end{itemize}

%% file: sec/2_related.tex
\input{table/motivation}

\section{Related Work}
\label{sec:related-work}
\subsection{Visual Grounding for Multimodal Reasoning}
Improving visual grounding has motivated interventions from both visual and language perspectives. Visual prompting methods \cite{yang2023set, yu2024attention,yang2023fine,khayatkhoei2025mllms} highlight informative regions with masks, spatial markers, and crops, helping VLMs focus on details in complex scenes.
Beyond static visual cues, guided search and learned visual operations \cite{wu2024v*,zheng2026deepeyes,zhang2025thyme,su2026pixel} enable models to inspect additional regions as reasoning unfolds.
From the language perspective, intermediate reasoning traces and reinforcement learning \cite{shao2024visual,huang2026vision,shen2025vlmr1,wang2026perception} promote structured, visually grounded predictions.
While these approaches highlight the value of evidence selection and structured reasoning, additional views or reasoning traces do not guarantee that visual evidence remains influential throughout generation. ReAccess instead makes previously processed evidence directly accessible within the model.

\subsection{Latent Memory and Test-Time Learning}
Recent work \cite{liu2025reasoningminddynamicmultimodal,semantic,pmlr-v306-jeon26c,wang2026monet,cheng2026hylar,tong2025sketch} explores latent representations for visual reasoning. LVR \cite{li2026latent} performs autoregressive visual reasoning by reconstructing visual tokens in latent space, while MemVR \cite{zou2024look} selectively reinjects encoded visual tokens into intermediate FFN layers to mitigate visual forgetting.
These advances highlight the importance of internal representations, motivating further exploration of how to organize and retrieve such information. Along this direction, test-time learning methods such as Titans \cite{behrouz2026titans} and $\delta$-mem \cite{lei2026deltamem} use associative memory to retain and retrieve long-context information. 
Building on these advances in long-term information management, ReAccess further explores the potential of test-time memory for complex visual reasoning, organizing visual evidence and directly integrating evidence retrieval into the model's reasoning.

%% file: table/motivation.tex
\begin{table*}[!t]

\caption{Visual evidence recoverability in compact representations.}
\vspace{-2.5mm}
\label{tab:motivation}

\centering
\small
\setlength{\tabcolsep}{9pt}
\renewcommand{\arraystretch}{0.70}
\begin{tabular}{@{}lrr rrrr@{}}
\toprule
 & \multicolumn{2}{c}{Baselines} & \multicolumn{4}{c}{Matched Image} \\
\cmidrule(lr){2-3}\cmidrule(lr){4-7}
Evidence Type & Question Only & Unrelated Image & Full State & PCA (16D) & PCA (64D) & Random (16D) \\
\midrule
Visual Appearance & 23.5 & 21.3 & 64.5 & 64.3 & \textbf{65.4} & 44.4 \\
Attribute Grounding & 51.5 & 53.5 & 85.9 & 86.3 & \textbf{86.6} & 84.3 \\
Spatial Structure & 53.2 & 51.0 & 89.5 & \textbf{90.5} & 89.8 & 78.8 \\
Physical Properties & 41.6 & 40.1 & \textbf{75.2} & 73.7 & 70.1 & 45.3 \\
\bottomrule
\end{tabular}
\vspace{-5mm}
\end{table*}

%% file: sec/3_motivation.tex
\section{Motivation}
\label{sec:motivation}

To investigate the gap between visual evidence representation and effective utilization, we examine three interconnected aspects: how relevant evidence is identified within complex scenes, how its accessibility changes during reasoning, and whether it can be preserved in compact internal representations using Qwen2.5-VL-7B \cite{Qwen2.5-VL} in \cref{fig:motivation}.

\noindent\textbf{Obs. 1: Focused evidence relies on global context.}
We first examine how a frozen VLM utilizes localized evidence across full images, oracle crops, and their combination. Though relevant regions occupy only 2.8\% of visual tokens, isolating them reduces accuracy, while adding them to the full image improves it (\cref{fig:motivation}(a)). This contrast highlights a key challenge in visual grounding: critical information often lies in small regions, yet its interpretation requires broader context. Notably, these gains cannot be attributed solely to finer visual detail. Adding tokens from unmagnified relevant visual tokens preserves most of the gains from enlarged crops (\cref{fig:motivation}(b)).
More notably, as the model's ability to attend to relevant regions diminishes, the gains from additional evidence become increasingly pronounced (\cref{fig:motivation}(c)). This trend further suggests that improving visual grounding requires not just finer details, but also effective access to critical evidence while preserving global context for its interpretation.

\noindent\textbf{Obs. 2: Explicit reasoning can weaken visual grounding.}
Identifying relevant evidence, however, does not guarantee its continued influence throughout generation. Since VLMs reason through language, explicitly describing visual observations offers a natural way to carry evidence into subsequent reasoning \cite{wei2022chain}. We examine this using a frozen VLM's native capabilities. Surprisingly, intermediate evidence descriptions and CoT increase computational cost without improving accuracy (\cref{fig:motivation}(d)). More revealingly, our attention analysis shows that the model progressively shifts its focus away from relevant image regions as textual reasoning unfolds, with larger declines for incorrect predictions (\cref{fig:motivation}(e)). Meanwhile, generated text increasingly competes with visual input for attention (\cref{fig:motivation}(f)).
These observations reveal a deeper limitation of explicit reasoning: visual evidence may lose influence as reliance on accumulated text grows. The challenge thus extends beyond preserving visual information to making relevant evidence effectively accessible when needed. We thus turn to the model's internal representations to assess what visual information remains available for reasoning.

\noindent\textbf{Obs. 3: Compact representations facilitate visual evidence access.}
We further investigate how task-relevant visual information is preserved in the hidden representations of a frozen VLM. Specifically, we apply dimensionality reduction to the last question-token representations \cite{chen2026last} and evaluate their ability to recover different types of visual evidence through linear probing. Remarkably, low-dimensional representations achieve comparable or even superior performance to the original high-dimensional states on several tasks (\cref{tab:motivation}). These findings suggest that rich representations contain substantial redundancy, while task-relevant visual information can remain readily accessible in a compact space. Beyond reducing storage and computational costs, such compactness provides a promising basis for more focused and efficient evidence retrieval, motivating a lightweight memory that connects evidence preservation with adaptive utilization throughout reasoning.

%% file: sec/4_method.tex
\begin{figure*}[t]
    \centering
    \includegraphics[width=\textwidth]{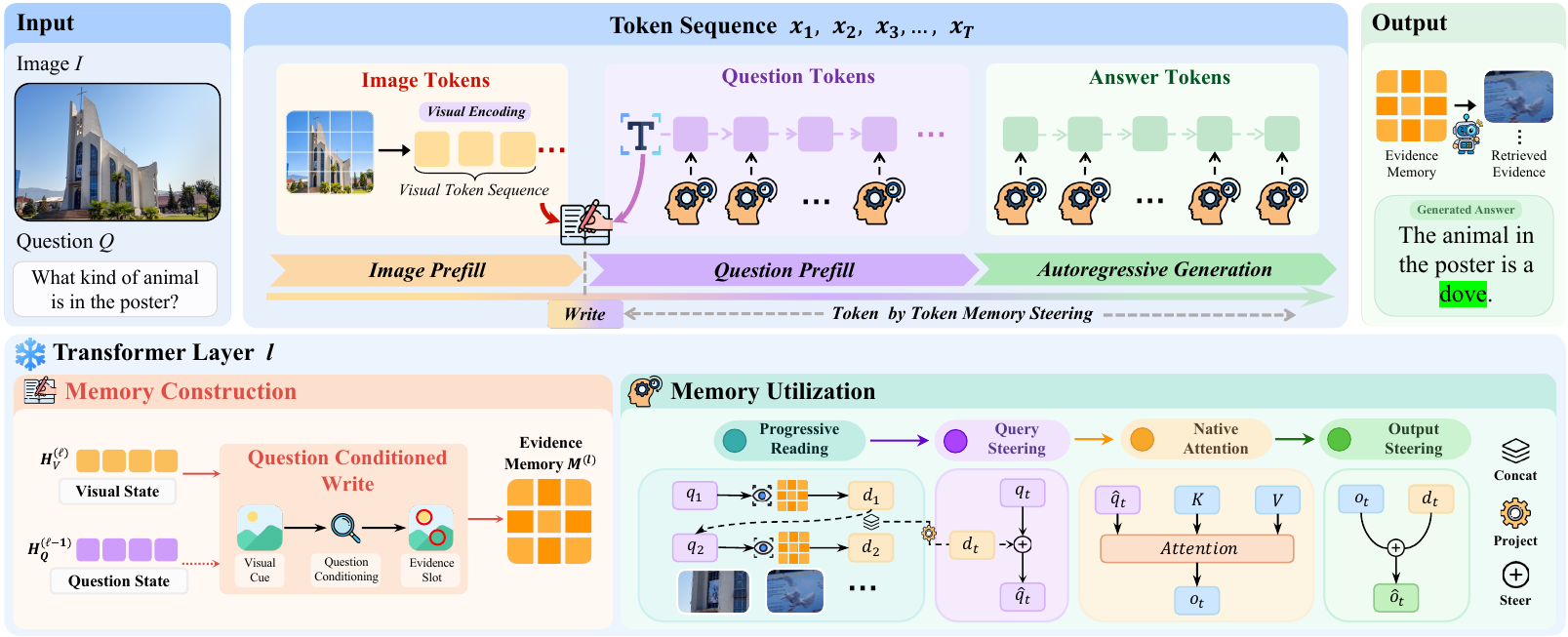}\par
    \vspace{-3mm}
\caption{
\textbf{Overview of ReAccess.}
Given an image--question pair, ReAccess constructs a compact associative memory from visual features during a single prefill (\textit{Memory Construction}). The memory is progressively accessed through successive reads to retrieve complementary visual evidence, which steers both attention queries and hidden representations of the frozen VLM (\textit{Memory Utility}).
}
\label{fig:framework}
\vspace{-6mm}
\end{figure*}
\section{Method}
\label{sec:method}

The analysis in \cref{sec:motivation} highlights two closely related challenges: identifying the visual evidence relevant to a question and ensuring that it continues to inform the answer. As illustrated in \cref{fig:framework}, \textbf{ReAccess} addresses both through a lightweight test-time evidence memory. \emph{Memory Construction} organizes question-relevant visual relationships into a compact state, while \emph{Memory Utility} recalls and incorporates them into the frozen VLM's native computation. The original visual pathway is preserved throughout.

\subsection{Preliminaries}
\label{sec:preliminaries}

Given an image $I$ and question $\mathcal Q$, let $f_\phi$ denote a frozen VLM with $L$ Transformer layers. At layer $\ell$, we denote the visual representations by $X_V^{(\ell)}=\{x_i^{(\ell)}\}_{i=1}^{N}$, the question representation by $z_{\mathcal Q}^{(\ell)}$, and the textual state at position $t$ by $x_t^{(\ell)}$. Following the standard associative-memory formulation \cite{schlag2021linear}, we use a compact matrix $S^{(\ell)}\in\mathbb R^{d\times d}$ to encode visual key--value associations, with a query $q$ retrieving content through $Sq$. Each layer constructs its memory during the original prefill for subsequent language states to access. We omit the layer index below when clear.

\subsection{Memory Construction}
\label{sec:memory-construction}

\noindent\textbf{Spatially aware evidence encoding.}
Fine-grained evidence often depends on both an object's appearance and its spatial surroundings. Preserving these local relationships is therefore important when organizing visual information for later retrieval. We project the visual features into compact key and value spaces through learnable mappings $W_K$ and $W_V$, and incorporate neighboring information using depthwise convolutions:
\begin{equation}
\label{eq:visual-associations}
\begin{aligned}[c]
k_i &= \operatorname{Normalize}\!\big(\operatorname{DWConv}_K(W_Kx_i)\big),\\
v_i &= \operatorname{Normalize}\!\big(\operatorname{DWConv}_V(W_Vx_i)\big).
\end{aligned}
\end{equation}
The keys provide a compact addressing space, while the values retain visual content informed by nearby regions, establishing the associations on which memory is built.

\noindent\textbf{Question-conditioned evidence selection.}
A scene offers many possible visual cues, but different questions require different evidence. Treating all associations equally can dilute the information most relevant to the answer. We therefore assign question-dependent weights to the spatially enriched visual keys:
\begin{equation}
\label{eq:question-weights}
\alpha_i=\operatorname{softmax}_i\!\left(
\frac{(W_s z_{\mathcal Q})^\top k_i}{\sqrt d}
\right).
\end{equation}
These soft weights emphasize informative regions while retaining contributions from the surrounding scene, reflecting the complementary role of local evidence and global context observed in \cref{sec:motivation}.

\noindent\textbf{Associative evidence consolidation.}
A weighted summary would merge distinct visual cues into a single representation, limiting their independent recovery as reasoning progresses. To preserve their associations for subsequent queries, we adopt a standard ridge-based \cite{hoerl1970ridge} write over the question-weighted keys and centered values. Defining $\bar v=\sum_i\alpha_i v_i$ and $C=\sum_i\alpha_i k_i k_i^\top$, we obtain
\begin{equation}
\label{eq:evidence-memory}
\begin{aligned}[c]
S &= \Big(\sum_i\alpha_i(v_i-\bar v)k_i^\top\Big)P,\\
P &= \Big(\tfrac{\operatorname{tr}(C)}{d}+\lambda_w\Big)
(C+\lambda_w\mathbf I_d)^{-1}.
\end{aligned}
\end{equation}
The regularized mapping stabilizes the associations among correlated visual features and retains a queryable representation of their differences. Its contents are further shaped by the evidence supervision described in \cref{sec:training-recipe}.

\subsection{Memory Utility}
\label{sec:memory-utility}
\noindent\textbf{Progressive evidence recall.}
Fine-grained visual evidence and its surrounding context provide complementary information for visual understanding. However, a single query derived from the current textual state may fail to capture their interdependencies. We therefore employ two sequential reads, allowing the initial readout to refine subsequent retrieval and progressively recover complementary evidence. With $d_{j,t}=\operatorname{RMSNorm}(Sq_{j,t})$ for $j\in\{1,2\}$, we compute
\begin{equation}
\label{eq:progressive-read}
\begin{aligned}[c]
q_{1,t} &= \operatorname{Normalize}(R_1x_t),\\
q_{2,t} &= \operatorname{Normalize}(R_2x_t+Ud_{1,t}),\\
r_t &= [d_{1,t};d_{2,t}].
\end{aligned}
\end{equation}
By conditioning subsequent retrieval on previously accessed evidence, this process facilitates the integration of local details and surrounding context, while retaining both representations for the current prediction.

\noindent\textbf{Evidence-guided reasoning.}
Retrieving relevant evidence is useful only when it influences how the VLM processes its existing context. We incorporate the readout through residual steering of the attention query and output. Let $\mathbf O_t(\widehat{\mathbf Q}_t)$ denote the native attention output evaluated using the corrected query. We compute
\begin{equation}
\label{eq:evidence-steering}
\begin{aligned}[c]
\widehat{\mathbf Q}_t
&=\mathbf Q_t+g_QW_{\Delta Q}r_t,\\
\widehat{\mathbf O}_t
&=\mathbf O_t(\widehat{\mathbf Q}_t)+g_OW_{\Delta O}r_t.
\end{aligned}
\end{equation}
The query correction guides access to the original visual and textual context; the output correction incorporates retrieved evidence into the updated representation. Together, they connect memory recall with the computations responsible for the answer.

\subsection{Training Recipe}
\label{sec:training-recipe}
We train ReAccess on VisCoT \cite{shao2024visual} with ground-truth evidence boxes, ensuring no train--test overlap. Lightweight memory modules are inserted across all layers of the frozen backbone. For Qwen2.5-VL-7B \cite{Qwen2.5-VL}, ReAccess requires only 58M trainable parameters (approximately 0.7\% of the full model) and one training epoch.

\noindent\textbf{Self-Guided Evidence Reconstruction.}
To encourage fine-grained evidence preservation, we introduce a self-guided reconstruction objective. The frozen VLM serves as its own teacher, generating concise evidence descriptions from ground-truth crops. An auxiliary training-time student branch then reconstructs these descriptions using only the question and full-image memory, without direct access to visual tokens. The reconstruction objective is 
\begin{equation}
\label{eq:evidence-loss}
\begin{aligned}[c]
\mathcal L_{\mathrm{ev}}
=-\sum_t \log p_{\phi,\theta}\big(
&y_t^{\mathrm{ev}}\mid y_{<t}^{\mathrm{ev}},
\mathcal Q,c_{\mathrm{ev}},\mathcal S_{I,\mathcal Q}
\big),
\end{aligned}
\end{equation}
where $y^{\mathrm{ev}}$ denotes the teacher-generated description and $c_{\mathrm{ev}}$ the evidence instruction. By reconstructing details directly accessible to the teacher but not the student, this objective encourages memory to retain fine-grained evidence.

\noindent\textbf{Spatially Grounded Retrieval.}
To align memory access with relevant visual regions, we derive a spatial retrieval distribution from the associative coefficients:
\begin{equation}
\label{eq:spatial-retrieval}
\pi_{i,t,j}^{(\ell)}=
\frac{\alpha_i^{(\ell)}
\left| (k_i^{(\ell)})^\top P^{(\ell)}q_{j,t}^{(\ell)}\right|}
{\sum_m\alpha_m^{(\ell)}
\left| (k_m^{(\ell)})^\top P^{(\ell)}q_{j,t}^{(\ell)}\right|}.
\end{equation}
Given the evidence-token set $\mathcal B$, the grounding objective maximizes retrieval mass within the annotated regions:

\begin{equation}
\label{eq:grounding-loss}
\begin{gathered}
\mu_{t,j}^{(\ell)}
= \sum_{i\in\mathcal B}\pi_{i,t,j}^{(\ell)},\\
\mathcal L_{\mathrm{ground}}
= -\mathbb E_{t\in\mathcal T_Q,j}
\left[
\log\left(
\frac{1}{L}\sum_{\ell=1}^{L}\mu_{t,j}^{(\ell)}
\right)\right].
\end{gathered}
\end{equation}
where $\mathcal T_Q$ denotes the question-token positions. This directly encourages evidence-grounded memory access without introducing a separate localization module.

\noindent\textbf{Complementary Evidence Recall.}
To prevent successive reads from retrieving redundant information, we encourage complementary evidence through a diversity objective:
\begin{equation}
\label{eq:diversity-loss}
\mathcal L_{\mathrm{div}}
=\mathbb E_{\ell,t,b}\left[
\cos^2\left(
\tilde d_{1,t}^{(\ell,b)},
\tilde d_{2,t}^{(\ell,b)}
\right)\right],
\end{equation}
where $\tilde d_{j,t}=Sq_{j,t}$ denotes the unnormalized readout and $b$ indexes the answer and evidence branches. Together with the answer-generation loss $\mathcal L_{\mathrm{ans}}$, the overall objective is
\begin{equation}
\label{eq:training-objective}
\begin{aligned}[c]
\mathcal L={}&\mathcal L_{\mathrm{ans}}
+\lambda_{\mathrm{ev}}\mathcal L_{\mathrm{ev}}
+\lambda_{\mathrm{ground}}\mathcal L_{\mathrm{ground}}
+\lambda_{\mathrm{div}}\mathcal L_{\mathrm{div}}.
\end{aligned}
\end{equation}
All annotations and descriptions are used exclusively during training. More details are provided in appendix.

%% file: sec/5_exp.tex
\input{table/main}

\input{table/baseline}
\input{table/ablation}
\input{table/memory_swap}
\section{Experiments}
\subsection{Experimental Setup}
\label{sec:experimental-setup}

\noindent\textbf{Benchmarks.}
We comprehensively evaluate ReAccess on diverse benchmarks covering general visual understanding, fine-grained perception, document understanding, and multimodal reasoning. These include GQA \cite{hudson2019gqa}, MMBench \cite{liu2024mmbench}, MMStar \cite{chen2024mmstar}, TextVQA \cite{singh2019textvqa}, DocVQA \cite{mathew2021docvqa}, InfoVQA \cite{mathew2022infographicvqa}, ChartQA \cite{masry2022chartqa}, AI2D \cite{kembhavi2016diagram}, OCRBench \cite{liu2024ocrbench}, MMMU \cite{yue2024mmmu}, LogicVista \cite{xiao2024logicvista}, and POPE \cite{pope}. We further evaluate challenging visual grounding and high-resolution perception using HR-Bench (4K/8K) \cite{wang2025hrbench}, V* \cite{wu2024v*}, and CV-Bench \cite{tong2024cvbench}. More details are provided in the Appendix.

\noindent\textbf{Baselines.}
We compare ReAccess against representative methods from four paradigms: (a) visual prompting methods, including SoM \cite{yang2023set}, API \cite{yu2024attention}, FGVP \cite{yang2023fine}, and ViCrop \cite{khayatkhoei2025mllms}; (b) active visual reasoning methods, including DeepEyes \cite{zheng2026deepeyes}, Thyme-RL \cite{zhang2025thyme}, and PixelReasoner \cite{su2026pixel}; (c) reasoning-oriented training methods, including Vision-R1 \cite{huang2026vision}, VLM-R1 \cite{shen2025vlmr1}, and PAPO \cite{wang2026perception}; and (d) latent reasoning methods including LVR \cite{li2026latent}, HyLaR \cite{cheng2026hylar}, SkiLa \cite{tong2025sketch} and Monet \cite{wang2026monet}. 

\noindent\textbf{Implementation Details.}
Unless otherwise specified, experiments use Qwen2.5-VL-7B \cite{Qwen2.5-VL} as the default backbone. The evidence memory is configured as a $64\times64$ matrix per Transformer layer. The image input pixel budget ranges from $256\times28\times28$ to $4096\times28\times28$ pixels. Further implementation details are provided in the Appendix.

\subsection{Main Results}
\noindent\textbf{Generalization Across VLM Backbones.}
As shown in \cref{tab:main_results}, ReAccess consistently enhances eight VLM backbones spanning diverse architectures and model scales across all twelve benchmarks. Gains are particularly pronounced on evidence-intensive benchmarks including MMStar, TextVQA, InfoVQA, and ChartQA, which require fine-grained visual details and relationships. Meanwhile, consistent gains on broader understanding and reasoning benchmarks also show that ReAccess complements the inherent capabilities of different backbones. With up to 16.6\% relative improvement, these results highlight the strong generalizability of ReAccess across visual understanding tasks.

\noindent\textbf{Comparison with Existing Methods.}
We further evaluate ReAccess on fine-grained visual perception and reasoning tasks, assessing generalization across input resolutions and visual granularity. As shown in \cref{tab:baseline_comparison}, ReAccess consistently improves upon the original model despite differing training and evaluation resolutions, achieving an average gain of 3.4 percentage points, with gains of up to 5.3 points. We group competing baselines by inference procedure: multi-turn, two-stage, and single-pass methods. Among them, ReAccess ranks among the top performers without fine-tuning the VLM backbone or relying on computationally expensive inference. Remarkably, it approaches the performance of multi-turn methods such as Thyme-RL and DeepEyes, even surpassing both on V*-Spatial and CV-Bench. These results demonstrate strong generalization to challenging visual scenarios while maintaining efficient inference.

\subsection{Understanding Evidence Memory}
\label{sec:memory-analysis}
Having established the effectiveness of ReAccess, We next examine how evidence memory drives these improvements.

\noindent\textbf{Component Ablations.}
We investigate how evidence preservation and utilization jointly contribute to ReAccess (\cref{tab:ablation}). Removing auxiliary supervision consistently degrades performance, with the absence of spatial grounding reducing HR-Bench 8K accuracy by 3.9 points. This suggests that preserving visual information alone is insufficient without reliable access to relevant evidence. More importantly, effective retrieval must translate into meaningful changes in model computation. Removing output steering, for instance, causes a substantial 9.2 points drop on V*-Spatial, while the remaining architectural ablations further highlight the importance of question-guided and progressive evidence integration. Together, these findings demonstrate that ReAccess benefits not merely from retaining visual evidence, but from making it accessible and actionable throughout reasoning.

\noindent\textbf{Memory Intervention.}
To assess the role of stored evidence, we investigate how the relevance and availability of visual information affect model performance.

\noindent\textit{Global Memory Replacement.}
We replace memory with variants constructed from an unrelated image or a mismatched question, alongside a zero-memory setting equivalent to the frozen VLM (\cref{tab:memory_swap}). Both reduce performance, highlighting the importance of task-aligned memory. Wrong-image memory even underperforms zero memory, suggesting that irrelevant evidence can impair predictions.

\begin{figure}[t]
    \centering
    \includegraphics[width=\columnwidth]{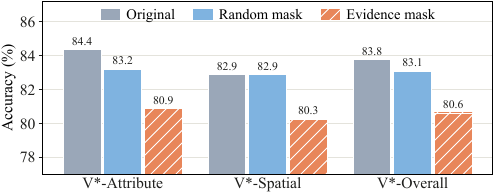}\par
    \vspace{-3mm}
    \caption{Causal evidence intervention on V*Bench.}
    \label{fig:casual}
    \vspace{-8mm}
\end{figure}

\noindent\textit{Regional Evidence Masking.}
To further test the model's reliance on specific visual evidence, we exclude ground-truth evidence regions during memory construction without altering the original image input (\cref{fig:casual}). Masking ground-truth evidence causes larger performance drops than random masking, especially for spatial reasoning. These results provide stronger causal evidence that ReAccess benefits from retaining and accessing task-relevant evidence rather than merely adding memory representations.

\begin{figure*}[!t]
    \centering
    \includegraphics[width=\textwidth]{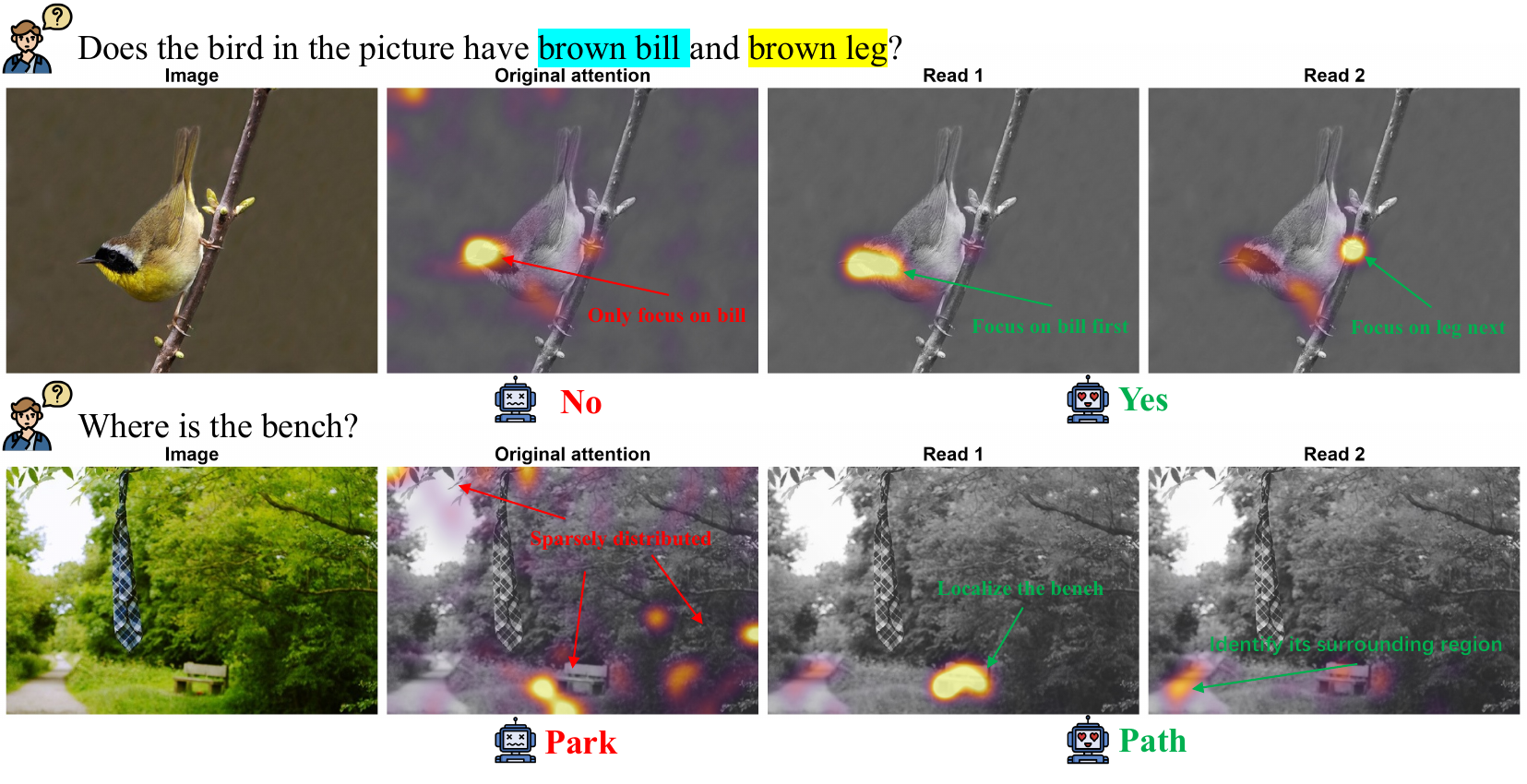}\par
    \vspace{-4mm}
\caption{Qualitative visualization of evidence preservation and utilization through ReAccess memory.}
    \label{fig:vis}
    \vspace{-5mm}
\end{figure*}

\noindent\textbf{Visualizing Evidence Utilization.}
To illustrate how ReAccess preserves and retrieves visual evidence from its evidence memory, \cref{fig:vis} contrasts original attention patterns with two successive memory readouts. The original model fails to adequately capture critical evidence, overlooking the bird's leg and struggling to identify the region surrounding the bench, leading to incorrect predictions. In contrast, ReAccess progressively retrieves complementary evidence from memory: successive reads focus on the bird's bill and leg, while the bench example proceeds from locating the object to identifying its surrounding path. These examples demonstrate how progressive evidence retrieval strengthens visual grounding and improves prediction accuracy.
\input{table/memory_size}
\begin{figure}[t]
    \centering
    \includegraphics[width=\columnwidth]{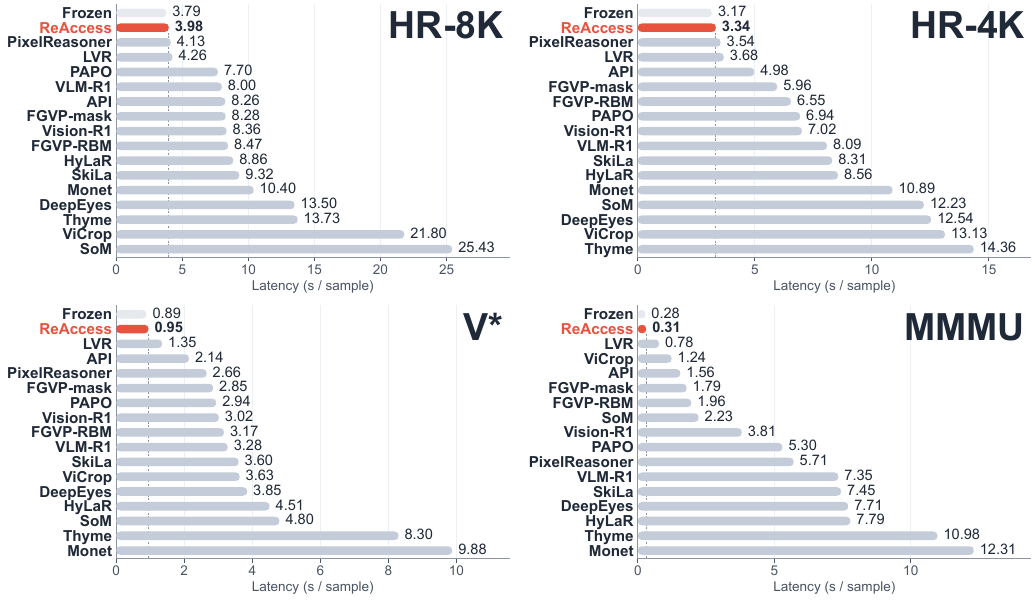}\par
    \vspace{-2.5mm}
    \caption{Inference efficiency comparison in different scenario}
    \label{fig:efficiency}
    \vspace{-7mm}
\end{figure}
\subsection{Designing Effective Evidence Memory}
\label{sec:efficiency}

\noindent\textbf{Inference Efficiency.}
Beyond improving visual understanding, ReAccess achieves a favorable balance between accuracy and computational efficiency. As shown in \cref{fig:efficiency}, ReAccess incurs only approximately 6\% additional latency over the frozen model on average, achieving the lowest inference latency among competing methods across diverse resolution settings. Unlike approaches that rely on task-dependent reasoning traces, ReAccess retrieves evidence directly from hidden states, maintaining a consistent procedure regardless of task complexity. Consequently, ReAccess achieves an average speedup of approximately 10$\times$ over methods such as Thyme-RL and DeepEyes, reaching up to 33$\times$ on reasoning-intensive benchmarks such as MMMU. These results demonstrate that ReAccess achieves strong visual understanding without costly explicit reasoning. Moreover, anomalous generations in competing methods can prolong inference and lead to incorrect answers. Even when these methods are evaluated solely on non-anomalous generations, ReAccess still achieves the lowest latency and exhibits no such anomalies (see appendix for further analysis).

\noindent\textbf{Memory Capacity.}
We next examine how memory capacity affects visual evidence preservation and utilization in \cref{tab:memory_size}.
Increasing memory size from $16 \times 16$ to $64 \times 64$ consistently improves performance, indicating that fine-grained evidence preservation requires sufficient capacity. However, further expanding it to $128 \times 128$ leads to performance degradation. This highlights the distinction between information retention and effective retrieval: larger memories may retain more visual detail, but can also introduce redundant or irrelevant associations competing with task-relevant evidence during retrieval. Consequently, greater capacity does not necessarily yield more useful evidence for reasoning. The $64 \times 64$ configuration best balances evidence preservation and retrieval selectivity, reinforcing the effectiveness of compact memory.

\noindent\textbf{Layer Insertion Depth.}
We further investigate the effect of memory insertion depth in \cref{tab:layer_depth}. Enabling memory access across all layers achieves the strongest performance, particularly on challenging fine-grained visual understanding tasks, suggesting that visual evidence is most effectively utilized when it can influence representations throughout the backbone. Among partial-layer configurations, middle-layer insertion performs best overall, achieving 80.3\% on V*-Spatial, compared with 75.0\% and 77.6\% for shallow and deep insertion, respectively. This suggests that intermediate layers offer a favorable balance between semantic abstraction and subsequent evidence integration. In contrast, shallow-layer representations may lack sufficient task-level context to guide evidence retrieval, while deep-layer insertion leaves limited opportunities for retrieved information to influence downstream computation. Full-depth insertion overcomes these limitations by allowing question-relevant evidence to be progressively incorporated as representations evolve.
\input{table/layer}
\section{Conclusion}
We presented ReAccess, a lightweight framework bridging visual evidence representation and utilization in VLMs. ReAccess enhances visual grounding by progressively integrating evidence from compact memory into native computation. With a single prefill, ReAccess improves visual understanding across diverse VLMs and benchmarks with minimal overhead. Further analyses show that these gains stem from accessing stored evidence rather than additional representations or computation. Our findings highlight that effective visual reasoning depends not only on perception, but also on retaining and accessing visual evidence, offering a promising direction for efficient multimodal reasoning.

%% file: table/main.tex
\definecolor{GainGreen}{HTML}{16825D}
\definecolor{OursTint}{HTML}{EDF2F8}
\definecolor{OursLabel}{HTML}{2F5275}
\definecolor{HeaderTint}{HTML}{F0F2F5}

\newcommand{\gain}[2]{%
  #1\,\textcolor{GainGreen}{\scriptsize$\uparrow$#2}%
}

\begin{table*}[t]
\centering
\caption{Performance on general visual understanding across various VLM backbones.}
\label{tab:main_results}
\vspace{-2.5mm}

\begingroup
\setlength{\tabcolsep}{1.9pt}
\renewcommand{\arraystretch}{0.85}

\resizebox{\textwidth}{!}{%
\begin{tabular}{@{}l *{12}{c}@{}}
\toprule[0.95pt]

\textbf{Backbone}
& \textbf{GQA}
& \textbf{MMBench}
& \textbf{MMStar}
& \textbf{TextVQA}
& \textbf{DocVQA}
& \textbf{InfoVQA}
& \textbf{ChartQA}
& \textbf{AI2D}
& \textbf{OCRBench}
& \textbf{MMMU}
& \textbf{LogicVista}
& \textbf{POPE} \\

\midrule[0.6pt]

\textbf{Qwen2.5-VL-7B} \cite{Qwen2.5-VL}
& 60.4 & 82.2 & 56.7 & 83.2
& 93.3 & 78.6 & 83.3 & 78.7
& 84.3 & 50.1 & 39.1 & 86.1 \\

\rowcolor{OursTint}
\textbf{\textcolor{OursLabel}{+Ours}}
& \gain{62.1}{1.7}
& \gain{83.7}{1.5}
& \gain{61.6}{4.9}
& \gain{85.0}{1.8}
& \gain{95.2}{1.9}
& \gain{80.5}{1.9}
& \gain{87.6}{4.3}
& \gain{82.6}{3.9}
& \gain{86.7}{2.4}
& \gain{51.0}{0.9}
& \gain{42.4}{3.3}
& \gain{87.9}{1.8} \\

\midrule[0.3pt]

\textbf{LLaVA-NeXT-7B} \cite{liu2023llava}
& 63.0 & 64.9 & 33.5 & 65.4
& 66.1 & 30.7 & 53.5 & 63.4
& 52.4 & 35.8 & 26.3 & 86.1 \\

\rowcolor{OursTint}
\textbf{\textcolor{OursLabel}{+Ours}}
& \gain{64.6}{1.6}
& \gain{67.4}{2.5}
& \gain{36.5}{3.0}
& \gain{67.8}{2.4}
& \gain{70.2}{4.1}
& \gain{35.8}{5.1}
& \gain{55.7}{2.2}
& \gain{65.2}{1.8}
& \gain{53.7}{1.3}
& \gain{37.4}{1.6}
& \gain{29.2}{2.9}
& \gain{88.2}{2.1} \\

\midrule[0.3pt]

\textbf{Qwen3-VL-8B} \cite{Qwen3-VL}
& 61.8 & 83.8 & 61.1 & 82.7
& 94.3 & 80.8 & 84.6 & 82.9
& 85.8 & 54.0 & 35.3 & 87.3 \\

\rowcolor{OursTint}
\textbf{\textcolor{OursLabel}{+Ours}}
& \gain{63.0}{1.2}
& \gain{86.4}{2.6}
& \gain{65.5}{4.4}
& \gain{84.6}{1.9}
& \gain{95.8}{1.5}
& \gain{83.6}{2.8}
& \gain{89.9}{5.3}
& \gain{84.3}{1.4}
& \gain{88.6}{2.8}
& \gain{55.7}{1.7}
& \gain{39.5}{4.2}
& \gain{88.5}{1.2} \\

\midrule[0.3pt]

\textbf{InternVL3.5-8B} \cite{wang2025internvl3}
& 58.3 & 81.1 & 64.6 & 74.6
& 90.2 & 73.6 & 84.5 & 81.0
& 83.7 & 58.4 & 37.3 & 89.7 \\

\rowcolor{OursTint}
\textbf{\textcolor{OursLabel}{+Ours}}
& \gain{61.1}{2.8}
& \gain{83.2}{2.1}
& \gain{66.6}{2.0}
& \gain{77.8}{3.2}
& \gain{92.5}{2.3}
& \gain{75.5}{1.9}
& \gain{87.5}{3.0}
& \gain{83.7}{2.7}
& \gain{86.0}{2.3}
& \gain{61.9}{3.5}
& \gain{40.0}{2.7}
& \gain{91.0}{1.3} \\

\midrule[0.3pt]

\textbf{Qwen2.5-VL-3B} \cite{Qwen2.5-VL}
& 54.6 & 77.8 & 55.8 & 79.9
& 93.0 & 74.2 & 82.5 & 76.2
& 79.7 & 44.2 & 32.8 & 86.5 \\

\rowcolor{OursTint}
\textbf{\textcolor{OursLabel}{+Ours}}
& \gain{56.7}{2.1}
& \gain{79.3}{1.5}
& \gain{58.4}{2.6}
& \gain{80.8}{0.9}
& \gain{93.2}{0.2}
& \gain{76.9}{2.7}
& \gain{83.8}{1.3}
& \gain{78.3}{2.1}
& \gain{82.7}{3.0}
& \gain{45.6}{1.4}
& \gain{34.4}{1.6}
& \gain{88.3}{1.8} \\

\midrule[0.3pt]

\textbf{LLaVA-OV-1.5-4B} \cite{an2025llava1.5}
& 55.7 & 82.1 & 60.5 & 76.8
& 90.4 & 70.1 & 84.5 & 82.2
& 78.7 & 51.7 & 39.1 & 86.0 \\

\rowcolor{OursTint}
\textbf{\textcolor{OursLabel}{+Ours}}
& \gain{57.0}{1.3}
& \gain{83.7}{1.6}
& \gain{63.6}{3.1}
& \gain{79.0}{2.2}
& \gain{92.1}{1.7}
& \gain{72.7}{2.6}
& \gain{86.1}{1.6}
& \gain{83.0}{0.8}
& \gain{81.2}{2.5}
& \gain{53.2}{1.5}
& \gain{42.4}{3.3}
& \gain{87.2}{1.2} \\

\midrule[0.3pt]

\textbf{LLaVA-OV-2-8B} \cite{an2026llavaov}
& 58.5 & 84.7 & 65.5 & 76.9
& 93.0 & 72.8 & 85.1 & 81.9
& 77.3 & 55.0 & 41.7 & 88.0 \\

\rowcolor{OursTint}
\textbf{\textcolor{OursLabel}{+Ours}}
& \gain{60.0}{1.5}
& \gain{86.7}{2.0}
& \gain{68.9}{3.4}
& \gain{79.7}{2.8}
& \gain{93.5}{0.5}
& \gain{74.3}{1.5}
& \gain{87.4}{2.3}
& \gain{83.5}{1.6}
& \gain{79.2}{1.9}
& \gain{56.7}{1.7}
& \gain{43.1}{1.4}
& \gain{89.4}{1.4} \\

\midrule[0.3pt]

\textbf{Gemma-3-12B} \cite{team2025gemma}
& 47.8 & 77.0 & 49.9 & 72.4
& 82.9 & 52.0 & 58.6 & 78.3
& 70.7 & 46.1 & 33.3 & 90.2 \\

\rowcolor{OursTint}
\textbf{\textcolor{OursLabel}{+Ours}}
& \gain{48.7}{0.9}
& \gain{78.8}{1.8}
& \gain{52.3}{2.4}
& \gain{77.7}{5.3}
& \gain{87.1}{4.2}
& \gain{57.2}{5.2}
& \gain{60.0}{1.4}
& \gain{80.8}{2.5}
& \gain{72.6}{1.9}
& \gain{48.3}{2.2}
& \gain{35.3}{2.0}
& \gain{90.7}{0.5} \\

\bottomrule[0.95pt]
\end{tabular}%
}

\endgroup
\vspace{-2mm}
\end{table*}

%% file: table/baseline.tex

\definecolor{NativeTint}{HTML}{F0F1F3}
\definecolor{NativeText}{HTML}{6F7680}
\definecolor{OursTint}{HTML}{EAF1F8}
\definecolor{OursText}{HTML}{24527A}

\newcommand{\native}[1]{\textcolor{NativeText}{#1}}

\begin{table*}[t]
\centering

\caption{Comparison with representative methods based on Qwen2.5-VL-7B. $\dagger$ denotes single-prefill variants.}

\label{tab:baseline_comparison}
\vspace{-2.5mm}

\begingroup
\small
\setlength{\tabcolsep}{10pt}
\renewcommand{\arraystretch}{0.4}

\begin{tabular}{lcccccc}
\toprule

\multirow{2}{*}{\textbf{Method}}
& \multirow{2}{*}{\textbf{HR-Bench 4K}}
& \multirow{2}{*}{\textbf{HR-Bench 8K}}
& \multicolumn{3}{c}{\textbf{V$^{*}$}}
& \multirow{2}{*}{\textbf{CV-Bench}} \\

\cmidrule(lr){4-6}
& & &
\textbf{Attribute}
& \textbf{Spatial}
& \textbf{Overall}
& \\

\midrule


\multicolumn{7}{c}{
\textit{Multi-turn}
} \\
\midrule

\rowcolor{NativeTint}
\native{Thyme-RL} \cite{zhang2025thyme}
& \native{77.0}
& \native{72.0}
& \native{84.4}
& \native{81.6}
& \native{83.3}
& \native{78.2} \\

\rowcolor{NativeTint}
\native{DeepEyes} \cite{zheng2026deepeyes}
& \native{73.9}
& \native{71.0}
& \native{88.7}
& \native{79.0}
& \native{84.8}
& \native{77.8} \\

\midrule


\multicolumn{7}{c}{
\textit{Two-stage}
} \\
\midrule

API \cite{yu2024attention}
& 71.5 & 68.5 & 80.9
& 75.0 & 78.5 & 78.1 \\

FGVP-Mask \cite{yang2023fine}
& 71.9 & 66.9 & 82.6
& 79.0 & 81.2 & 77.2 \\

FGVP-RBM \cite{yang2023fine}
& 41.1 & 42.6 & 27.8
& 50.0 & 36.7 & 52.3 \\

SoM \cite{yang2023set}
& 51.6 & 41.4 & 33.0
& 65.8 & 46.1 & 50.1 \\

ViCrop \cite{khayatkhoei2025mllms}
& 72.1 & 67.9 & 83.5
& 72.4 & 79.1 & 76.4 \\

\midrule


\multicolumn{7}{c}{
\textit{Single-pass}
} \\
\midrule

Qwen2.5-VL-7B \cite{Qwen2.5-VL}
& 70.9 & 66.9 & 81.7
& 77.6 & 80.1 & 77.7 \\

Vision-R1 \cite{huang2026vision}
& 72.5 & 70.9 & 80.0
& 80.3 & 80.1 & 75.4 \\

VLM-R1 \cite{shen2025vlmr1}
& 64.5 & 61.6 & 80.9
& 61.8 & 73.3 & 64.2 \\

LVR \cite{li2026latent}
& 67.5 & 64.6 & 83.5
& 76.3 & 80.6 & 77.7 \\


HyLaR \cite{cheng2026hylar}
& 73.1 & 70.4 & 83.5
& 75.0 & 80.1 & 75.7 \\

SkiLa \cite{tong2025sketch}
& 72.3 & 64.6 & 84.4
& 81.6 & 83.3 & 76.9 \\

Monet \cite{wang2026monet}
& 69.0 & 65.4 & 83.5
& 82.9 & 83.3 & 66.3 \\


PAPO \cite{wang2026perception}
& 72.0 & 69.1 & 81.7
& 79.0 & 80.6 & 78.2 \\

PixelReasoner \cite{su2026pixel}
& 72.3 & 65.9 & 80.9
& 79.0 & 80.1 & 77.9 \\

Thyme-RL$^{\dagger}$ \cite{zhang2025thyme}
& 72.3 & 67.8 & 81.7
& 80.3 & 81.2 & 77.5 \\

DeepEyes$^{\dagger}$ \cite{zheng2026deepeyes}
& 71.0 & 68.3 & 80.9
& 75.0 & 78.5 & 76.3 \\


\rowcolor{OursTint}
\textbf{\textcolor{OursText}{ReAccess (Ours)}}
& \textbf{73.3}
& \textbf{71.8}
& \textbf{84.4}
& \textbf{82.9}
& \textbf{83.8}
& \textbf{78.8} \\

\bottomrule
\end{tabular}

\endgroup
\vspace{-5mm}
\end{table*}

%% file: table/ablation.tex
\providecommand{\abYes}{\textcolor{gray!65}{\ding{51}}}
\providecommand{\abNo}{\textcolor{red!65!black}{\ding{55}}}
\definecolor{AblationTint}{HTML}{EDF2F8}

\begin{table*}[t]
\centering
\caption{Component Ablations of ReAccess.}
\label{tab:ablation}
\vspace{-2.5mm}
\begingroup
\small
\setlength{\tabcolsep}{3.0pt}
\renewcommand{\arraystretch}{1.04}
\resizebox{\textwidth}{!}{%
\begin{tabular}{@{}l *{3}{>{\centering\arraybackslash}m{1.15cm}} cccc ccccccc@{}}
\toprule
 & \multicolumn{3}{c}{\textbf{Training Objectives}}
 & \multicolumn{4}{c}{\textbf{Architecture}}
 & \multicolumn{7}{c}{\textbf{Evaluation Benchmarks}} \\
\cmidrule(lr){2-4}\cmidrule(lr){5-8}\cmidrule(l){9-15}
\textbf{Configuration} & $\mathcal{L}_{\mathrm{ev}}$ & $\mathcal{L}_{\mathrm{ground}}$ & $\mathcal{L}_{\mathrm{div}}$
& \textbf{Q} & \textbf{O} & \textbf{Q-Cond.} & \textbf{Prog.}
& \textbf{HR-4K} & \textbf{HR-8K}
& \textbf{V$^*$-Attr.} & \textbf{V$^*$-Spat.} & \textbf{V$^*$-Overall}
& \textbf{TextVQA} & \textbf{POPE} \\
\midrule
\rowcolor{AblationTint}
\textbf{Full} & \abYes & \abYes & \abYes & \abYes & \abYes & \abYes & \abYes
& \textbf{73.3} & \textbf{71.8} & \textbf{84.4} & \textbf{82.9} & \textbf{83.8} & \textbf{85.0} & \textbf{87.9} \\
\midrule[0.4pt]
\multicolumn{15}{@{}l}{\textit{Training objective ablations}} \\
\midrule[0.35pt]
Evidence reconstruction & \abNo & \abYes & \abYes & \multicolumn{4}{c}{}
& 71.6 & 69.8 & 82.6 & 79.0 & 81.2 & 83.4 & 87.2 \\
Grounding & \abYes & \abNo & \abYes & \multicolumn{4}{c}{}
& 72.0 & 67.9 & 80.9 & 77.6 & 79.6 & 84.2 & 86.7 \\
Diversity & \abYes & \abYes & \abNo & \multicolumn{4}{c}{}
& 72.4 & 70.8 & 83.5 & 80.3 & 82.2 & 84.6 & 87.3 \\
\midrule[0.4pt]
\multicolumn{15}{@{}l}{\textit{Architecture ablations}} \\
\midrule[0.35pt]
Query steering & \multicolumn{3}{c}{} & \abNo & \abYes & \abYes & \abYes
& 72.8 & 70.5 & 83.5 & 77.6 & 81.2 & 84.6 & 86.8 \\
Output steering & \multicolumn{3}{c}{} & \abYes & \abNo & \abYes & \abYes
& 73.0 & 70.3 & 81.7 & 73.7 & 78.5 & 84.7 & 87.6 \\
Question conditioning & \multicolumn{3}{c}{} & \abYes & \abYes & \abNo & \abYes
& 72.1 & 69.1 & 84.4 & 76.3 & 81.2 & 84.0 & 87.0 \\
Progressive reading & \multicolumn{3}{c}{} & \abYes & \abYes & \abYes & \abNo
& 72.5 & 71.3 & 82.6 & 81.6 & 82.2 & 84.6 & 87.2 \\
\bottomrule
\end{tabular}%
}
\endgroup
\vspace{-5mm}
\end{table*}

%% file: table/memory_swap.tex
\definecolor{MemoryTint}{HTML}{EDF2F8}

\begin{table}[t]
\centering
\caption{Assessing the Contribution of Evidence Memory}
\label{tab:memory_swap}
\vspace{-2.5mm}
\small
\setlength{\tabcolsep}{4pt}
\renewcommand{\arraystretch}{0.92}

\resizebox{\columnwidth}{!}{%
\begin{tabular}{@{}lccccc@{}}
\toprule
& \multicolumn{2}{c}{\textbf{HR-Bench}}
& \multicolumn{3}{c}{\textbf{V*}} \\
\cmidrule(lr){2-3}
\cmidrule(l){4-6}
\textbf{Memory Setting}
& \textbf{4K}
& \textbf{8K}
& \textbf{Attr.}
& \textbf{Spatial}
& \textbf{Overall} \\
\midrule
\rowcolor{MemoryTint}
\textbf{Original}
& \textbf{73.3}
& \textbf{71.8}
& \textbf{84.4}
& \textbf{82.9}
& \textbf{83.8} \\
Wrong-Image
& 69.6 & 66.1 & 80.0 & 76.3 & 78.5 \\
Wrong-Question
& 71.3 & 68.4 & 82.6 & 79.0 & 81.2 \\
Zero-Memory
& 70.9 & 66.9 & 81.7 & 77.6 & 80.1 \\
\bottomrule
\end{tabular}%
}
\vspace{-3.5mm}
\end{table}

%% file: table/memory_size.tex

\definecolor{MemoryTint}{HTML}{EDF2F8}

\begin{table}[t]
\centering
\caption{Effect of memory size on visual understanding.}
\label{tab:memory_size}
\vspace{-2mm}

\begingroup
\small
\setlength{\tabcolsep}{3.5pt}
\renewcommand{\arraystretch}{1.22}

\resizebox{\columnwidth}{!}{%
\begin{tabular}{@{}lcccccc@{}}
\toprule

\multirow{2}{*}{\textbf{Memory Size}}
& \multicolumn{2}{c}{\textbf{HR-Bench}}
& \multicolumn{3}{c}{\textbf{V*}}
& \multirow{2}{*}{\textbf{TextVQA}} \\

\cmidrule(lr){2-3}
\cmidrule(lr){4-6}

& \textbf{4K}
& \textbf{8K}
& \textbf{Attr.}
& \textbf{Spatial}
& \textbf{Overall}
& \\

\midrule

$128 \times 128$
& 71.8
& 69.8
& 81.7
& 79.0
& 80.6
& 84.7 \\

\rowcolor{MemoryTint}
$\mathbf{64 \times 64}\ (default)$
& \textbf{73.3}
& \textbf{71.8}
& \textbf{84.4}
& \textbf{82.9}
& \textbf{83.8}
& \textbf{85.0} \\

$32 \times 32$
& 73.1
& 71.5
& 83.5
& 81.6
& 82.7
& 84.6 \\

$16 \times 16$
& 72.8
& 71.0
& 82.6
& 81.6
& 82.2
& 84.1 \\

\bottomrule
\end{tabular}%
}

\endgroup
\vspace{-6mm}
\end{table}

%% file: table/layer.tex

\definecolor{DepthTint}{HTML}{EDF2F8}

\begin{table}[t]
\centering
\caption{Effect of memory insertion depth on performance.}
\label{tab:layer_depth}
\vspace{-2mm}

\begingroup
\small
\setlength{\tabcolsep}{3.5pt}
\renewcommand{\arraystretch}{0.90}

\resizebox{\columnwidth}{!}{%
\begin{tabular}{@{}lcccccc@{}}
\toprule

\multirow{2}{*}{\textbf{Depth}}
& \multicolumn{2}{c}{\textbf{HR-Bench}}
& \multicolumn{3}{c}{\textbf{V*}}
& \multirow{2}{*}{\textbf{TextVQA}} \\

\cmidrule(lr){2-3}
\cmidrule(lr){4-6}

& \textbf{4K}
& \textbf{8K}
& \textbf{Attr.}
& \textbf{Spatial}
& \textbf{Overall}
& \\

\midrule

\rowcolor{DepthTint}
\textbf{Full}
& \textbf{73.3}
& \textbf{71.8}
& \textbf{84.4}
& \textbf{82.9}
& \textbf{83.8}
& \textbf{85.0} \\

Shallow
& 72.0
& 68.9
& 80.9
& 75.0
& 78.5
& 84.6 \\

Middle
& 72.5
& 69.0
& 83.5
& 80.3
& 82.2
& 84.0 \\

Deep
& 72.1
& 68.8
& 81.7
& 77.6
& 80.1
& 82.9 \\

\bottomrule
\end{tabular}%
}

\endgroup
\vspace{-6.5mm}
\end{table}

%% file: sec/X_suppl.tex
\clearpage
\setcounter{page}{1}
\maketitlesupplementary

\noindent\textbf{Overview.}
This appendix provides additional details and analyses to complement the main paper. We first elaborate on the implementation of ReAccess, including mathematical derivations, training objectives, and hyperparameter configurations. We then provide details for the evaluated VLM backbones and benchmarks, followed by a validation of the robustness and stability of our proposed method.


\section{Implementation Details}
\label{app:implementation}

\subsection{Detailed Memory Formulation}
\label{app:memory_formulation}

\noindent\textbf{Visual Feature Preparation.}
ReAccess constructs an independent associative memory at each language Transformer layer. Given visual representations $\{x_i\}_{i=1}^{N}$ with hidden dimension $H$, learnable projections transform them into a compact space of dimension $d$. We further incorporate local spatial context through depthwise convolutions:
\begin{equation}
\begin{aligned}
k_i &=
\operatorname{unit}\left(
\operatorname{DWConv}_K(W_Kx_i)
\right),\\
v_i &=
\operatorname{RMSNorm}\left(
\operatorname{DWConv}_V(W_Vx_i)
\right),
\end{aligned}
\end{equation}
where $k_i,v_i\in\mathbb R^d$, and $\operatorname{unit}(\cdot)$ denotes $\ell_2$ normalization.

The question representation $z_{\mathcal Q}$ is formed by averaging the textual representations preceding the assistant response, using token embeddings at the first layer. Question-dependent weights are then computed as
\begin{equation}
\alpha_i =
\operatorname{softmax}_i
\left(
\frac{(W_sz_{\mathcal Q})^\top k_i}{\sqrt d}
\right).
\end{equation}
These weights determine the relative contribution of visual associations during memory construction.

\noindent\textbf{Closed-Form Associative Writing.}
We interpret memory construction as a regularized regression problem over the question-weighted visual associations. To distinguish individual visual features from their shared component, we first center the values using their weighted mean:
\begin{equation}
\bar v=\sum_{i=1}^{N}\alpha_i v_i.
\end{equation}
We define the weighted second-moment matrix and the key-value association matrix as
\begin{equation}
\label{eq:app_memory_statistics}
\begin{aligned}
C &= \sum_{i=1}^{N}\alpha_i k_i k_i^\top,\\
B &= \sum_{i=1}^{N}
\alpha_i(v_i-\bar v)k_i^\top.
\end{aligned}
\end{equation}

The corresponding weighted ridge-regression objective is
\begin{equation}
\label{eq:app_ridge_objective}
\mathcal J(A)=
\sum_{i=1}^{N}
\alpha_i
\left\|
Ak_i-(v_i-\bar v)
\right\|_2^2
+
\lambda\|A\|_F^2,
\end{equation}
where $\lambda>0$ controls regularization.

\noindent\textbf{Proposition (Scaled Ridge Solution).}
For fixed visual features and question-dependent weights, the memory constructed by ReAccess is a scaled solution of the weighted ridge-regression objective in \cref{eq:app_ridge_objective}.

\noindent\textit{Proof.}
Expanding the objective and differentiating with respect to $A$ yields
\begin{equation}
\begin{aligned}
\nabla_A\mathcal J
&=
2\sum_i
\alpha_i
\left(Ak_i-(v_i-\bar v)\right)k_i^\top
+2\lambda A\\
&=2(AC-B+\lambda A).
\end{aligned}
\end{equation}
Setting the gradient to zero gives
\begin{equation}
A(C+\lambda I)=B.
\end{equation}
Since $C$ is positive semidefinite and $\lambda>0$, the matrix $C+\lambda I$ is positive definite and therefore invertible. The unique solution is
\begin{equation}
\label{eq:app_ridge_solution}
A^*=B(C+\lambda I)^{-1}.
\end{equation}

ReAccess applies an additional scale calibration to this solution. Specifically, defining
\begin{equation}
\label{eq:app_preconditioner}
c=\frac{\operatorname{tr}(C)}{d}+\lambda,
\qquad
P=c(C+\lambda I)^{-1},
\end{equation}
the resulting memory is
\begin{equation}
\label{eq:app_memory_solution}
\boxed{
S=BP=cA^*.
}
\end{equation}
Thus, the constructed memory is a scaled ridge-regression solution specialized to the current visual representations and question. \hfill$\square$

This formulation provides a closed-form test-time adaptation mechanism without gradient-based updates. Each layer stores a fixed $d\times d$ matrix, whose size is independent of the number of visual tokens. After feature projection, constructing the memory requires $O(Nd^2+d^3)$ operations for matrix accumulation and inversion.

\noindent\textbf{Progressive Memory Retrieval.}
Given a textual hidden state $x_t$, ReAccess performs two sequential memory reads:
\begin{equation}
\label{eq:app_progressive_reading}
\begin{aligned}
q_{1,t}
&=\operatorname{unit}(R_1x_t),\\
\tilde d_{1,t}
&=Sq_{1,t},\\
d_{1,t}
&=\operatorname{RMSNorm}(\tilde d_{1,t}),\\
q_{2,t}
&=\operatorname{unit}(R_2x_t+Ud_{1,t}),\\
\tilde d_{2,t}
&=Sq_{2,t},\\
d_{2,t}
&=\operatorname{RMSNorm}(\tilde d_{2,t}).
\end{aligned}
\end{equation}
The second query incorporates information retrieved by the first read, allowing subsequent access to depend on previously retrieved evidence. Both readouts are retained through concatenation:
\begin{equation}
r_t=[d_{1,t};d_{2,t}]\in\mathbb R^{2d}.
\end{equation}

Substituting \cref{eq:app_memory_solution} into the retrieval operation gives
\begin{equation}
\label{eq:app_memory_decomposition}
Sq_{j,t}
=
\sum_{i=1}^{N}
\underbrace{
\alpha_i(k_i^\top Pq_{j,t})
}_{\text{retrieval coefficient}}
(v_i-\bar v).
\end{equation}
This decomposition distinguishes question-dependent association weights from the state-dependent retrieval coefficients. Importantly, the decomposition is analytical: inference accesses the stored matrix directly without recomputing the contribution of every visual token.

\noindent\textbf{Integration with Native Computation.}
The concatenated readout modulates the native attention query and output through
\begin{equation}
\label{eq:app_memory_injection}
\begin{aligned}
\mathbf Q'_t
&=\mathbf Q_t+g_QW_{\Delta Q}r_t,\\
\mathbf O'_t
&=\mathbf O_t+g_OW_{\Delta O}r_t,
\end{aligned}
\end{equation}
where
\begin{equation}
g_Q=\sigma(a_Q),
\qquad
g_O=\sigma(a_O).
\end{equation}
The query correction is applied before multimodal rotary positional encoding (mRoPE). The native attention output $\mathbf O_t$ is computed using the corrected query and is subsequently modified after the output projection.

The original attention and KV-cache mechanisms remain in place. Memory construction occurs once per layer during the original prefill, and the resulting state remains fixed during autoregressive decoding. In contrast, retrieval queries and memory-induced corrections vary with the evolving textual representations.

\subsection{Training Objectives and Supervision}
\label{app:training_objectives}

\noindent\textbf{Evidence Reconstruction.}
To encourage the associative memory to preserve question-relevant visual information, we employ privileged evidence supervision during training. The frozen Qwen2.5-VL-7B backbone acts as a teacher, receiving the question and an evidence crop extracted from the VisCoT-annotated bounding box, with a crop pixel budget ranging from
$256 \times 28 \times 28$ to $512 \times 28 \times 28$ pixels. Operating along its native visual pathway without memory augmentation or access to the reference answer, the teacher generates a concise evidence description using greedy decoding with a maximum of 32 tokens.

The student receives the question and the memory constructed from the complete image, but no direct image tokens. It reconstructs the teacher-generated description under teacher forcing. Both teacher generation and student reconstruction use the same evidence supervision prompt:

\begin{tcolorbox}[
    colback=blue!3,
    colframe=blue!45!black,
    boxrule=0.6pt,
    arc=1.5mm,
    left=2mm,
    right=2mm,
    top=2mm,
    bottom=2mm
]
{\footnotesize\ttfamily
\{Question\}\par
Provide evidence to answer the question in one sentence.
}
\end{tcolorbox}

The reconstruction objective is defined as
\begin{equation}
\label{eq:app_evidence_sft}
\mathcal L_{\mathrm{SFT}}
=
-\sum_{t=1}^{|y^{\mathrm{ev}}|}
\log p_{\phi,\theta}
\left(
y_t^{\mathrm{ev}}
\mid
y_{<t}^{\mathrm{ev}},
\mathcal Q,
c_{\mathrm{ev}},
\mathcal S_{I,\mathcal Q}
\right),
\end{equation}
where $y^{\mathrm{ev}}$ denotes the teacher-generated evidence description and $c_{\mathrm{ev}}$ is the shared evidence instruction.

Because direct visual tokens are excluded from the student branch, image-conditioned information must be accessed through the constructed memory. This provides explicit supervision for evidence recoverability without requiring evidence descriptions during inference.

\noindent\textbf{Spatial Grounding.}
Although memory retrieval is conditioned on the language state, the retrieved associations are not necessarily aligned with the visual evidence supporting the question. We therefore introduce spatial grounding supervision based on the decomposition in \cref{eq:app_memory_decomposition}.

For each retrieval query, we define the magnitude-based contribution of visual token $i$ as
\begin{equation}
\label{eq:app_retrieval_weight}
w_{i,t,j}^{(\ell)}
=
\alpha_i^{(\ell)}
\left|
(k_i^{(\ell)})^\top
P^{(\ell)}q_{j,t}^{(\ell)}
\right|.
\end{equation}
The corresponding normalized spatial distribution is
\begin{equation}
\label{eq:app_retrieval_distribution}
\pi_{i,t,j}^{(\ell)}
=
\frac{
w_{i,t,j}^{(\ell)}
}{
\sum_m w_{m,t,j}^{(\ell)}
}.
\end{equation}

Let $\mathcal B^{(\ell)}$ denote the visual tokens corresponding to the annotated evidence region at layer $\ell$. The retrieval mass assigned to this region is
\begin{equation}
\mu_{t,j}^{(\ell)}
=
\sum_{i\in\mathcal B^{(\ell)}}
\pi_{i,t,j}^{(\ell)}.
\end{equation}
We further aggregate the evidence mass across memory layers:
\begin{equation}
\label{eq:app_grounding_mass}
\bar\mu_{t,j}
=
\frac{1}{L}
\sum_{\ell=1}^{L}
\mu_{t,j}^{(\ell)}.
\end{equation}

The grounding objective penalizes the negative logarithm of the layer-aggregated evidence mass, using the question positions and both memory reads. This directly constrains the memory retrieval process without introducing a separate visual localization head.

The spatial distribution reflects the magnitude of memory retrieval coefficients and serves as a proxy for evidence access, rather than a direct causal attribution of the model's final prediction.

\noindent\textbf{Complementary Retrieval.}
Sequential retrieval alone does not prevent the two readouts from producing redundant representations. We therefore encourage complementary retrieval by penalizing their squared cosine similarity:
\begin{equation}
\label{eq:app_diversity_loss}
\mathcal L_{\mathrm{div}}
=
\mathbb E_{\ell,t,b}
\left[
\frac{
\left(
\tilde d_{1,t}^{(\ell,b)\top}
\tilde d_{2,t}^{(\ell,b)}
\right)^2
}{
\|\tilde d_{1,t}^{(\ell,b)}\|_2^2
\|\tilde d_{2,t}^{(\ell,b)}\|_2^2
}
\right],
\end{equation}
where $\tilde d_{j,t}$ denotes the readout before RMS normalization, and $b$ indexes the answer and evidence-reconstruction branches. This objective is applied across all memory layers and applicable reader positions.

\noindent\textbf{Joint Optimization.}
The standard answer objective supervises answer tokens and the end-of-response token:
\begin{equation}
\mathcal L_{\mathrm{answer}}
=
-\sum_{t\in\mathcal Y}
\log p_{\phi,\theta}
\left(
y_t
\mid
I,\mathcal Q,y_{<t};
\mathcal S_{I,\mathcal Q}
\right),
\end{equation}
where $\mathcal Y$ denotes the supervised output positions.

The four objectives are optimized jointly:
\begin{equation}
\label{eq:app_joint_objective}
\begin{aligned}
\mathcal L
={}&
\mathcal L_{\mathrm{answer}}
+0.3\mathcal L_{\mathrm{SFT}}\\
&+0.2\mathcal L_{\mathrm{ground}}
+0.05\mathcal L_{\mathrm{div}}.
\end{aligned}
\end{equation}
All losses contribute to the memory modules within a single optimization step, while the backbone remains frozen. The evidence teacher is used only to generate training targets and does not participate in inference.

\subsection{Training Configurations}
\label{app:training_configurations}

We train ReAccess on the VisCoT training split using four NVIDIA A800 80GB GPUs with distributed data parallelism. The training procedure uses 396,256 examples for one epoch, corresponding to 6,192 optimizer updates with a global batch size of 64.

Only the lightweight memory modules are optimized, comprising 57,949,780 trainable parameters (approximately 0.7\% of the backbone) in total. All 28 language Transformer layers are equipped with independent memory parameters.

We use AdamW with separate learning rates and weight decay for matrix projections, scalar parameters, and spatial convolutions. A linear warmup is followed by cosine learning-rate decay. The backbone operates in BF16, while memory parameters, construction, retrieval, and associated optimization are maintained in FP32.

Training examples are grouped into micro-batches according to their visual-token lengths to limit padding overhead, while gradient accumulation maintains the specified global batch size. Detailed configurations are provided in \cref{tab:app_hyperparameters}.
\input{table/appendix/hp}

\section{Datasets and Evaluation}
\label{app:datasets_evaluation}

\subsection{Vision-Language Backbones}
\label{app:backbones}

To evaluate the generalizability of ReAccess, we conduct experiments across seven vision-language backbones with different model scales and architectural designs. All backbone parameters remain frozen, and ReAccess is integrated through lightweight, trainable memory modules.

\noindent\textbf{Qwen2.5-VL-3B.}
Qwen2.5-VL-3B is a compact vision-language model from the Qwen2.5-VL family. It employs a vision Transformer with dynamic-resolution image processing and multimodal rotary positional embeddings (mRoPE), enabling the model to accommodate images with varying spatial resolutions. We include this model to examine whether ReAccess can improve visual understanding under a relatively limited model capacity.

\noindent\textbf{Qwen2.5-VL-7B.}
Qwen2.5-VL-7B is a larger variant of the same model family, retaining its dynamic-resolution visual encoding and multimodal positional modeling. It provides stronger capabilities in visual perception, document understanding, and multimodal reasoning. We adopt this backbone for our primary experiments, including comparisons with existing methods, component ablations, and detailed memory analyses.

\noindent\textbf{Qwen3-VL-8B.}
Qwen3-VL-8B is a newer-generation multimodal model with enhanced visual perception and reasoning capabilities. Its architecture incorporates improved multimodal positional encoding and multi-level visual feature integration through DeepStack, allowing visual information from different depths of the vision encoder to contribute to language modeling. This backbone evaluates the applicability of ReAccess to more recent multimodal architectures.

\noindent\textbf{InternVL3.5-8B.}
InternVL3.5-8B belongs to the InternVL family of multimodal models, combining a dedicated vision encoder with a large language model through visual feature alignment. It supports dynamic visual processing and is designed for general image understanding, document interpretation, and multimodal reasoning. We use this backbone to assess the compatibility of ReAccess with an alternative vision-language architecture.

\noindent\textbf{LLaVA-NeXT-7B.}
LLaVA-NeXT-7B extends the LLaVA architecture with dynamic high-resolution image processing and improved visual instruction tuning. Its AnyRes strategy represents images through multiple visual patches, preserving additional spatial details compared with conventional fixed-resolution inputs. This backbone allows us to investigate whether ReAccess remains effective when visual information is already processed through a high-resolution representation strategy.

\noindent\textbf{LLaVA-OneVision-1.5-4B.}
LLaVA-OneVision-1.5-4B is a compact multimodal model built upon a Rice vision Transformer and a Qwen3 language backbone. It supports native-resolution image processing and employs spatial token merging for efficient visual representation. We include this model to assess the generalizability of ReAccess across different visual architectures and multimodal integration designs.

\noindent\textbf{LLaVA-OneVision-2-8B.}
LLaVA-OneVision-2-8B is a recent model from the LLaVA-OneVision series, designed to support unified image and video understanding. It introduces a native OneVision visual encoder with windowed attention for efficient visual feature processing. We include this model to examine the generalizability of ReAccess to a different visual encoding and multimodal integration design.

\noindent\textbf{Gemma-3-12B.}
Gemma-3-12B is a multimodal model from the Gemma family, integrating a SigLIP-based vision encoder with a language model. Its visual processing operates on fixed-resolution image inputs, with optional adaptive cropping through a pan-and-scan mechanism for high-resolution images. As the largest backbone in our evaluation, it further examines the scalability of ReAccess across model sizes and visual processing architectures.
\subsection{Evaluation Benchmarks}
\label{app:benchmarks}

We evaluate ReAccess on twelve general visual understanding benchmarks and three additional benchmarks targeting fine-grained visual perception and reasoning. Together, these datasets cover object recognition, visual reasoning, scene text, document understanding, spatial relationships, and high-resolution perception.

\noindent\textbf{GQA.}
GQA evaluates compositional visual question answering in natural images. Its questions are constructed around scene graphs and require understanding object attributes, relationships, and multi-step visual dependencies. The benchmark emphasizes compositional reasoning while reducing biases associated with question-answer distributions. Performance is measured using answer accuracy.

\noindent\textbf{MMBench.}
MMBench is a comprehensive multiple-choice benchmark designed to evaluate general multimodal understanding. It covers a diverse range of perception and reasoning abilities, including object recognition, fine-grained perception, attribute reasoning, and relational understanding. Its evaluation framework incorporates circular evaluation to reduce sensitivity to answer-option ordering, providing a more robust assessment of multimodal capabilities.

\noindent\textbf{MMStar.}
MMStar is a vision-dependent multimodal benchmark designed to assess whether models genuinely use visual information when answering questions. It contains 1,500 carefully curated multiple-choice samples spanning six major dimensions, including perception, instance reasoning, logical reasoning, science and technology, and mathematics. The benchmark emphasizes challenging visual reasoning while reducing the influence of language-only shortcuts. Performance is evaluated using multiple-choice accuracy.

\noindent\textbf{TextVQA.}
TextVQA evaluates the ability to read and reason about textual information embedded in natural scene images. Questions require recognizing scene text and integrating it with surrounding visual context, such as identifying signs, labels, or written information associated with objects. The dataset contains over 45,000 questions and uses the standard VQA accuracy metric.

\noindent\textbf{DocVQA.}
DocVQA evaluates visual question answering over document images, including forms, reports, and other structured documents. Answering its questions requires recognizing textual content while interpreting document layouts and the spatial relationships between different elements. Performance is measured using Average Normalized Levenshtein Similarity (ANLS), which accounts for minor textual differences between predicted and reference answers.

\noindent\textbf{InfoVQA.}
InfoVQA focuses on question answering over infographic images, where information is conveyed through combinations of text, illustrations, charts, and spatial layouts. Models must integrate textual recognition with visual structure to identify relevant information and produce accurate answers. The benchmark assesses the ability to interpret visually complex, information-dense content and is evaluated using ANLS.

\noindent\textbf{ChartQA.}
ChartQA evaluates question answering over data visualizations, including bar charts, line charts, and pie charts. Its questions require understanding chart structures, extracting numerical values, comparing quantities, and performing basic arithmetic or logical reasoning. Performance is evaluated using relaxed accuracy, allowing limited numerical variation in predicted answers.

\noindent\textbf{AI2D.}
AI2D evaluates understanding and reasoning over scientific and educational diagrams. The dataset contains illustrations with objects, labels, arrows, and other structural elements that encode scientific concepts or relationships. Answering its multiple-choice questions requires interpreting both visual components and their semantic connections. Performance is reported using answer accuracy.

\noindent\textbf{OCRBench.}
OCRBench provides a comprehensive evaluation of text recognition and text-oriented visual understanding in multimodal models. It includes tasks such as scene text recognition, text-centric visual question answering, document-oriented question answering, key information extraction, and handwritten mathematical expression recognition. The benchmark measures whether models can accurately recover and utilize textual evidence across diverse visual settings, using its official aggregate scoring protocol.

\noindent\textbf{MMMU.}
MMMU evaluates advanced multimodal understanding and reasoning across academic and professional disciplines. It contains questions involving diagrams, charts, scientific illustrations, and other specialized visual materials, requiring domain knowledge together with visual interpretation. The benchmark spans multiple disciplines and includes both multiple-choice and open-ended questions. Performance is measured using the official answer accuracy protocol.

\noindent\textbf{LogicVista.}
LogicVista evaluates fundamental logical reasoning in visual contexts. It contains 448 questions covering five reasoning categories: deductive, inductive, numerical, spatial, and mechanical reasoning. These tasks involve interpreting visual patterns, diagrams, spatial configurations, and logical relationships rather than relying solely on object recognition. Performance is evaluated through multiple-choice accuracy.

\noindent\textbf{POPE.}
POPE evaluates object hallucination in vision-language models through binary questions about object presence. It probes whether a model incorrectly identifies objects that are absent from an image, using random, popular, and adversarial negative sampling strategies. The benchmark assesses the reliability of visual grounding and the tendency to produce unsupported object claims.

\noindent\textbf{HR-Bench.}
HR-Bench evaluates fine-grained perception and reasoning over high-resolution images. It provides separate 4K and 8K evaluation settings, where visually relevant details may occupy only a small portion of the complete image. Questions require models to identify subtle attributes or relationships while retaining sufficient global context. The benchmark is particularly relevant for evaluating whether ReAccess improves access to localized evidence in visually complex scenes.

\noindent\textbf{V*.}
V* evaluates fine-grained visual understanding in high-resolution and visually crowded images. It comprises two categories: Attribute Recognition, which requires identifying detailed properties of target objects, and Spatial Relationship Reasoning, which examines relationships between objects in complex scenes. These tasks require precise localization and interpretation of potentially small or visually inconspicuous regions. We report performance on both categories and their overall accuracy.

\noindent\textbf{CV-Bench.}
CV-Bench is a vision-centric benchmark designed to evaluate fundamental two-dimensional and three-dimensional visual understanding. Its 2D tasks include object counting and spatial relationship recognition, while its 3D tasks assess depth ordering and relative distance estimation. The benchmark contains 2,638 manually inspected questions derived from established computer vision datasets. It examines whether multimodal models can perform basic visual perception and spatial reasoning beyond general semantic understanding.
\input{table/appendix/latency}

\input{table/appendix/failure}
\subsection{Evaluation Protocol}
\label{app:evaluation_protocol}

All models are evaluated following their official inference configurations, including the corresponding image preprocessing and conversation formatting procedures. We adopt the evaluation protocols and scoring metrics specified by each benchmark, ensuring consistent comparisons between the original backbones and ReAccess. No additional evidence crops, external visual tools, or intermediate evidence generation are introduced during ReAccess inference.
\subsection{Comparison Baselines}
\label{app:baselines}

We evaluate all baseline methods following their respective official inference protocols and configurations. Below, we briefly introduce each baseline.

\noindent\textbf{Thyme-RL.}
Thyme-RL enables multimodal reasoning through executable code that performs image manipulation and computational operations. It combines supervised fine-tuning with reinforcement learning to develop adaptive tool-use capabilities for fine-grained perception and complex visual reasoning.

\noindent\textbf{DeepEyes.}
DeepEyes introduces a thinking-with-images framework that learns to actively inspect visual evidence through reinforcement learning. By leveraging the model's inherent grounding capabilities, it performs iterative region localization and visual cropping without relying on specialized external perception models.

\noindent\textbf{PixelReasoner.}
PixelReasoner extends visual reasoning into pixel space by incorporating operations such as image zooming and frame selection. It employs instruction tuning followed by curiosity-driven reinforcement learning to encourage effective exploration of visual evidence.

\noindent\textbf{LVR.}
Latent Visual Reasoning (LVR) enables autoregressive reasoning directly in the continuous visual embedding space. It generates intermediate latent states that reconstruct task-relevant visual representations, allowing visual and textual reasoning to interact during generation.

\noindent\textbf{Vision-R1.}
Vision-R1 enhances multimodal reasoning through a two-stage training framework combining cold-start supervised fine-tuning and reinforcement learning. It employs progressively constrained reasoning trajectories to improve reasoning accuracy while mitigating excessive textual deliberation.

\noindent\textbf{VLM-R1.}
VLM-R1 extends R1-style reinforcement learning to vision-language models through rule-based rewards. It leverages verifiable task-specific feedback to improve visual grounding and multimodal reasoning without requiring learned reward models.

\noindent\textbf{API.}
Attention Prompting on Image (API) employs an auxiliary vision-language model to generate question-guided attention maps. These maps are overlaid on the original image to emphasize relevant visual regions and guide subsequent perception.

\noindent\textbf{FGVP-Mask.}
Fine-Grained Visual Prompting (FGVP) uses segmentation-based visual cues to highlight target objects. Its Mask variant emphasizes the selected region through pixel-level masking, reducing interference from irrelevant visual content.

\noindent\textbf{FGVP-RBM.}
FGVP-RBM adopts Blur Reverse Mask prompting, which blurs image regions outside the target segmentation mask. This preserves contextual spatial information while emphasizing the visual details of the target object.

\noindent\textbf{SoM.}
Set-of-Mark (SoM) prompting overlays numbered markers on segmented image regions to establish explicit visual references. These markers allow vision-language models to associate textual descriptions with specific objects and spatial locations.

\noindent\textbf{ViCrop.}
ViCrop is a training-free visual cropping approach that identifies question-relevant image regions using internal attention or gradient information. It introduces magnified regional observations alongside the original image to recover fine-grained visual details.

\noindent\textbf{PAPO.}
Perception-Aware Policy Optimization (PAPO) improves multimodal reinforcement learning by explicitly encouraging visual grounding. It introduces a perception-sensitive KL objective contrasting original and corrupted visual inputs, together with entropy regularization for stable optimization.

\noindent\textbf{HyLaR.}
Hybrid Latent Reasoning (HyLaR) interleaves textual reasoning with continuous visual latent representations through a Canvas-in-Latents mechanism. It further introduces Decoupled Policy Optimization to jointly optimize discrete textual and continuous latent reasoning.

\noindent\textbf{SkiLa.}
Sketch-in-Latents (SkiLa) enables multimodal models to alternate between explicit textual reasoning and continuous latent visual sketches. A semantic reconstruction objective guides the generation of visual thoughts, supporting interleaved reasoning without explicit image reconstruction.

\noindent\textbf{Monet.}
Monet enables reasoning through continuous latent visual embeddings that serve as intermediate visual thoughts. It combines a multi-stage distillation-based training pipeline with Visual-latent Policy Optimization to strengthen latent visual reasoning.
\section{Inference Stability}
\label{app:stability}

To characterize typical inference efficiency, the main paper reports latency after excluding anomalous generations that exhaust the maximum new-token limit, preventing extreme outliers from obscuring general inference trends. However, such anomalies also reflect an important aspect of practical inference stability. We therefore provide a complementary analysis of unfiltered inference latency in this appendix. Specifically, we randomly sample 100 examples from each of four benchmarks: HR-Bench 4K, HR-Bench 8K, V*, and MMMU, yielding 400 evaluation samples per method. We measure end-to-end inference latency, retaining all samples, including anomalous generations and retries, to capture their actual computational costs.

As shown in \cref{tab:stability_latency}, ReAccess exhibits **no anomalous generations across all 400 samples**, matching the stability of the frozen backbone while introducing only marginal latency overhead. In contrast, several competing methods exhibit substantial instability, with 29 anomalous cases for Vision-R1, 20 for Monet, 10 for HyLaR, and 6 for Thyme-RL, alongside additional failures in other methods. These anomalies are particularly prevalent on MMMU, where prolonged reasoning and repeated generation can substantially inflate inference costs. As detailed in \cref{tab:stability_failure_modes}, representative failures involve repetitive outputs, exhausted generation limits, and repeated inference attempts, resulting in unpredictable latency spikes. Unlike approaches relying on potentially lengthy, task-dependent reasoning traces, ReAccess directly retrieves visual evidence from hidden states through a consistent inference procedure, avoiding such generation-related overhead in our evaluation. These results highlight that ReAccess not only improves fine-grained visual understanding efficiently, but also preserves the runtime predictability and inference stability of the original backbone, making its computational cost more reliable across diverse tasks.

%% file: table/appendix/hp.tex

\begin{table}[t]
\centering
\caption{\textbf{Architecture and training hyperparameters of ReAccess.}}
\label{tab:app_hyperparameters}
\vspace{0.5mm}

\small
\setlength{\tabcolsep}{4pt}
\renewcommand{\arraystretch}{1.10}

\begin{tabularx}{\columnwidth}{
@{}l l >{\raggedright\arraybackslash}X@{}
}
\toprule

\textbf{Configuration}
& \textbf{Parameter}
& \textbf{Value} \\

\midrule

\multirow{6}{*}{Core}
& Backbone
& Qwen2.5-VL-7B \\

& Memory dimension
& 64 \\

& Memory layers
& 28 \\

& Memory reads
& 2 \\

& Spatial kernel
& Depthwise $3\times3$ \\

& Trainable parameters
& 57.9M ($\sim$0.7\%) \\

\midrule

\multirow{10}{*}{Training}
& Dataset
& VisCoT \\

& Batch size
& 64 \\

& Epochs
& 1 \\

& Visual tokens
& 256--4,096 \\

& Learning rate
& $2\times10^{-4}$ / $2\times10^{-3}$ \\

& Warmup ratio
& 0.03 \\

& Optimizer
& AdamW \\

& Scheduler
& Cosine \\

& Gradient clipping
& 1.0 \\

& Loss weights
& 0.3 / 0.2 / 0.05 \\

\bottomrule
\end{tabularx}
\end{table}

%% file: table/appendix/latency.tex

\definecolor{StabilityTint}{HTML}{EDF2F8}
\definecolor{StabilityGray}{HTML}{777E89}

\providecommand{\lat}[2]{%
  #1\,{\color{StabilityGray}\scriptsize(#2)}%
}

\begin{table*}[t]
\centering
\caption{\textbf{Inference latency and anomalous generations across benchmarks.}
Average latency (s/sample) includes anomalous generations and retries. Parentheses indicate anomalous sample counts.}
\label{tab:stability_latency}
\vspace{0.5mm}

\small
\setlength{\tabcolsep}{7pt}
\renewcommand{\arraystretch}{1.08}

\begin{tabularx}{\textwidth}{
@{}l*{5}{>{\centering\arraybackslash}X}@{}
}
\toprule

& \multicolumn{2}{c}{\textbf{HR-Bench}}
& \textbf{V*}
& \textbf{MMMU}
& \textbf{Anomalies} \\

\cmidrule(lr){2-3}
\cmidrule(lr){4-4}
\cmidrule(lr){5-5}
\cmidrule(l){6-6}

\textbf{Method}
& \textbf{8K}
& \textbf{4K}
& \textbf{Latency}
& \textbf{Latency}
& \textbf{/400} \\

\midrule

\rowcolor{StabilityTint}
\textbf{ReAccess}
& \textbf{\lat{3.98}{0}}
& \textbf{\lat{3.34}{0}}
& \textbf{\lat{0.95}{0}}
& \textbf{\lat{0.31}{0}}
& \textbf{0} \\

Frozen
& \lat{3.79}{0}
& \lat{3.17}{0}
& \lat{0.89}{0}
& \lat{0.28}{0}
& 0 \\

\midrule[0.35pt]

LVR
& \lat{4.26}{0}
& \lat{3.78}{0}
& \lat{1.38}{0}
& \lat{1.00}{0}
& 0 \\

API
& \lat{8.26}{0}
& \lat{4.98}{0}
& \lat{2.14}{0}
& \lat{1.24}{0}
& 0 \\

FGVP-Mask
& \lat{8.28}{0}
& \lat{5.93}{0}
& \lat{2.86}{0}
& \lat{1.70}{0}
& 0 \\

FGVP-RBM
& \lat{8.47}{0}
& \lat{6.55}{0}
& \lat{3.02}{0}
& \lat{1.94}{0}
& 0 \\

SoM
& \lat{25.43}{0}
& \lat{12.23}{0}
& \lat{4.80}{0}
& \lat{2.23}{0}
& 0 \\

ViCrop
& \lat{21.80}{0}
& \lat{13.13}{0}
& \lat{3.63}{0}
& \lat{5.30}{0}
& 0 \\

VLM-R1
& \lat{8.60}{0}
& \lat{8.09}{0}
& \lat{3.17}{0}
& \lat{7.35}{0}
& 0 \\

PAPO
& \lat{7.70}{0}
& \lat{6.84}{0}
& \lat{2.94}{0}
& \lat{1.56}{0}
& 0 \\

\midrule[0.35pt]

PixelReasoner
& \lat{4.13}{0}
& \lat{3.54}{0}
& \lat{2.66}{0}
& \lat{5.71}{1}
& 1 \\

Vision-R1
& \lat{11.93}{7}
& \lat{9.30}{3}
& \lat{3.02}{0}
& \lat{32.94}{19}
& 29 \\

DeepEyes
& \lat{13.50}{0}
& \lat{12.54}{0}
& \lat{3.85}{0}
& \lat{16.71}{2}
& 2 \\

Thyme-RL
& \lat{13.73}{0}
& \lat{17.36}{1}
& \lat{8.30}{0}
& \lat{22.78}{5}
& 6 \\

HyLaR
& \lat{9.15}{1}
& \lat{8.56}{0}
& \lat{4.51}{0}
& \lat{9.76}{9}
& 10 \\

SkiLa
& \lat{9.32}{0}
& \lat{8.31}{0}
& \lat{3.60}{0}
& \lat{7.87}{2}
& 2 \\

Monet
& \lat{10.40}{0}
& \lat{11.71}{2}
& \lat{10.31}{1}
& \lat{19.06}{17}
& 20 \\

\bottomrule
\end{tabularx}
\end{table*}

%% file: table/appendix/failure.tex
\begin{table}[t]
\centering
\caption{\textbf{Causes of anomalous inference latency.}
Anomalies primarily arise from repetitive generation and retry mechanisms.}
\label{tab:stability_failure_modes}
\vspace{0.5mm}

\small
\setlength{\tabcolsep}{5pt}
\renewcommand{\arraystretch}{1.12}

\begin{tabularx}{\columnwidth}{
@{}l c >{\raggedright\arraybackslash}X@{}
}
\toprule
\textbf{Method}
& \textbf{Anomalies}
& \textbf{Failure Cause} \\
\midrule

Thyme-RL
& 6
& Repetitive outputs reaching token limits and triggering repeated retries. \\

DeepEyes
& 2
& Repeated unsuccessful action requests with excessive token generation. \\

Vision-R1
& 29
& Overlong, repetitive reasoning and retry loops during sampled decoding. \\

PixelReasoner
& 1
& Repetitive generation exhausting the output token limit. \\

HyLaR
& 10
& Erroneous generation continuing until the maximum token limit. \\

SkiLa
& 2
& Erroneous generation continuing until the maximum token limit. \\

Monet
& 20
& Erroneous generation continuing until the maximum token limit. \\

\bottomrule
\end{tabularx}
\end{table}